\RequirePackage{amsmath}
\documentclass[runningheads]{llncs}
\usepackage{graphicx}
\usepackage{mathtools}
\usepackage{multirow}
\usepackage{color}
\usepackage{hyperref}

\usepackage[tight,footnotesize]{subfigure}
\usepackage{epstopdf}
\AppendGraphicsExtensions{.tif}

\usepackage[open]{bookmark}

\begin{document}
\title{SafeDrones: Real-Time Reliability Evaluation of UAVs using Executable Digital Dependable Identities}
%
\titlerunning{SafeDrones}
\author{Koorosh Aslansefat\inst{1} 
\and Panagiota Nikolaou\inst{2} 
\and Martin Walker \inst{1} 
\and Mohammed Naveed Akram\inst{3} 
\and Ioannis Sorokos\inst{3} 
\and Jan Reich \inst{3} 
\and Panayiotis Kolios \inst{2}
\and Maria K. Michael \inst{2}
\and Theocharis Theocharides \inst{2}
\and Georgios Ellinas \inst{2}
\and Daniel Schneider\inst{3} 
\and Yiannis Papadopoulos\inst{1} }
\authorrunning{K. Aslansefat et al.}
\institute{University of Hull, United Kingdom 
\and KIOS Research and Innovation Center Of Excellence \& Department of Electrical and Computer Engineering, University Of Cyprus, Nicosia, Cyprus 
\and Fraunhofer Institute for Experimental Software Engineering (IESE), Germany
}

\maketitle              
\begin{abstract}
The use of Unmanned Arial Vehicles (UAVs) offers many advantages across a variety of applications. However, safety assurance is a key barrier to widespread usage, especially given the unpredictable operational and environmental factors experienced by UAVs, which are hard to capture solely at design-time. 
This paper proposes a new reliability modeling approach called SafeDrones to help address this issue by enabling runtime reliability and risk assessment of UAVs. It is a prototype instantiation of the Executable Digital Dependable Identity (EDDI) concept, which aims to create a model-based solution for real-time, data-driven dependability assurance for multi-robot systems. By providing real-time reliability estimates, SafeDrones allows UAVs to update their missions accordingly in an adaptive manner. 

\end{abstract}
\section{Introduction}
\label{sec_intro}
\pdfbookmark[section]{Introduction}{sec_intro}
There are many potential applications for Unmanned Aerial Vehicles (UAVs), including logistics, emergency response, filming, traffic monitoring, search and rescue, rail surveillance, and infrastructure inspection \cite{belcastro2017hazards}. However, one of the major barriers to widespread deployment and acceptance of UAVs is that of safety, particularly for operations in urban areas where UAV failure brings a higher risk of harm. For instance, during testing for Amazon's planned fast drone-based delivery service, their drones crashed five times over a four-month period in 2021 \cite{Amazon_Air_Bloomberg}. 
Therefore, safety and reliability must be key objectives during both the design and operation of UAVs to help minimise risk and improve likelihood of mission success \cite{sadeghzadeh2011fault}. 


Reliability can be defined generally as the probability of a system functioning correctly over a given period of time \cite{trivedi2017reliability}. There is a long history of reliability engineering techniques intended to help analyse, understand, and prevent failures. Among the most popular are Fault Tree Analysis (FTA) \cite{Vesely2002} and Failure Modes and Effects Analysis (FMEA). Such techniques were originally manually applied but over time have evolved and now form integral parts of comprehensive, tool-supported methodologies, encompassed under Model-Based Safety Analysis \cite{sharvia2016}. Using such approaches during the design of a UAV, it is possible to determine the ways in which it can fail and the likelihood of those failures.

Even so, UAVs must often operate independently in dynamic, unpredictable environments with varying mission goals, all of which are difficult to capture in a design-time analysis model. By combining design-time knowledge with safety monitoring applied at runtime, we can perform dynamic reliability evaluation and obtain a clearer picture of UAV reliability during operation.

This paper proposes a new model-based approach to improve reliability and safety of UAVs called SafeDrones. SafeDrones builds upon static design-time knowledge in the form of fault trees by combining them with dynamic Markov-based models and real-time monitoring to perform continuous reliability evaluation at runtime. The result is a modular safety monitor known as an Executable Dependable Digital Identity (EDDI), which can then be used to inform operational decision making.

The rest of the paper is organized as follows: 
in section 2, a brief background on relevant techniques is provided. Section 3 introduces SafeDrones and explains its proposed methodology. An experimental implementation is described in section 4 and results discussed in section 5, to illustrate the capabilities and limitations of the idea. Finally, conclusions and future research directions are presented.

\section{Background}
\label{Background}
\pdfbookmark[section]{Background}{sec_background}

\subsection{Fault Tree Analysis}
\label{sec_fta}
\pdfbookmark[subsection]{Fault Tree Analysis}{sec_fta}
Fault Tree Analysis (FTA) is a widely used, top-down reasoning approach for reliability analysis. It begins with a given system failure (the top event) and progressively explores possible combinations of causes until individual component failures have been reached. It is capable of both qualitative (logical) and quantitative (probabilistic) analysis, but requires the use of extensions to model dynamic relationships between causes and components and thus model dynamic failure behaviour \cite{aslansefat2020}. Markov chains are often used to perform the quantitative analysis of dynamic fault trees but this can result in significant complexity due to the state-space explosion problem \cite{bouissou2003,adler2007}. 

Although these limitations make fault trees less common for runtime applications, fault trees have also been used at runtime, particularly for diagnosis purposes. For example, by connecting sensor readings as inputs to "complex basic events" (CBE) in the fault tree, the failure model can be updated in real-time to provide potential diagnoses of problems, predictions of future failures, and reliability evaluation \cite{Complex_BE}. Such an approach is extended and used in SafeDrones.

There has also been some work on the use of FTA for UAVs specifically. Reliability improvement of UAVs through FTA and FMEA in design review procedures has been studied in \cite{franco2007failure}. \cite{murtha2009evidence} proposes a procedure for reliability improvement in a cost-effective way based on FTA. It also reduced the uncertainty in failure data via Dempster-Shafer theory. In \cite{olson2013qualitative}, the Michigan UAV is evaluated through qualitative failure analysis. However, these papers address only the design-time evaluation of UAVs, while in SafeDrones, both design-time and runtime evaluation are assessed.

\subsection{Reliability Modelling using Semi-Markov Processes (SMP)} 
\label{SMPbasedREL}
\pdfbookmark[subsection]{Reliability Modelling using Semi-Markov Processes (SMP)}{SMPbasedREL}
Given the limitations of FTA when applied to highly dynamic systems, particularly adaptive, unmanned systems like UAVs, they are often combined with other techniques such as Markov Processes and Petri-Nets. Markov Processes are a well-known tool for reliability evaluation, particularly when it comes to dynamic behaviours such as repair, priority, and sequences. A Semi-Markov Process (SMP) is a special type of Markov process that has the ability to work with non-exponential failure distributions \cite{trivedi2017reliability}. SMP can be defined using the following set: $(p, P, F(t))$. $p$ is the initial probability distribution vector, $P$ is the conditional transition probabilities matrix, and $F(t)$ describes the matrix of distribution functions of sojourn times in $i^{th}$ state, when $j^{th}$ state is next. 

Let $X_i$, $\forall_{i}= 0, 1, 2, \ldots$ be random variables. The time-homogeneous SMP $X$ is determined by a vector of initial state probabilities $p(0)$ and the matrix of conditional transition probability $P(t)=\lceil P_{ij}(t)\rceil$ is computed by Eq. \eqref{eqn1}. In this equation, $P_{ij}(t)$ satisfies the Kolmogorov-Feller equation \cite{trivedi2017reliability}.
\begin{equation}\label{eqn1}
P_{ij}(t)=P\{X(t)=j| X(0)=i\}=\delta_{ij}[1-G_{i}(t)]+\mathop{\sum_{k\in S}\mathop{\int_{0}^{t}P_{kj}}(t-x)dQ_{ik}(x)}
\end{equation}

where $\delta_{ij}=1$ if $i=j$  and $\delta_{ij}=0$  otherwise, $G_i$  is the distribution of the sojourn time in state $i$ described by Eq. \eqref{eqn5} \cite{cochran2010wiley}, and $Q_{ij}(t)$ describes the kernel matrix by Eq. \eqref{eqn5}.    

\begin{equation}\label{eqn5}
    G_{i}(t)=P\{S_{i}\leq t\mid X_{0}=i\}=\sum_{j=1}^{i}Q_{ij}(t)
\end{equation}

where $S_{i}, i=0,1,2,\ldots$  is the state of the system at time t and $Q_{ij}(t)$ is $P\{X_{1}=j,S_{i}\leq t\mid X_{0}=i\}$. The solution of Eq. \eqref{eqn1} can be found by applying the Laplace Stieltjes Transformation (LST) in matrix form by Eq. \eqref{eqn9} \cite{aslansefat2019hierarchical}. Note that $\tilde{p}(t)$ represents the probability vector in time domain and $\tilde{p}(s)$ represents the probability vector in LST domain.



\begin{equation}\label{eqn9}
    \tilde{p}(s)={\lceil 1- \tilde{q}(s)\rceil}^{-1}\tilde{g}(s))
\end{equation}

Having solved Eq. \eqref{eqn9} with taking the inverse LST of $\tilde{p}(s)$, the unconditional state probabilities in time domain are determined as $P(t)=P(0)P(t)$. Finally, the reliability of a system can be computed through summation of probability of operational state in the SMP. 
To calculate the MTTF, the transition matrix should have the following canonical form where there are r absorbing states (forming R) and t transient states (forming Q).

\begin{equation}\label{eqn12}
P=\left[\begin{matrix}Q&R\\0&I\\\end{matrix}\right]
\end{equation}

For the fundamental $P$ matrix, consider $N={(I-Q)}^{-1}$  and let $t_i$ be the average number of steps before reaching the absorbing state, given that the chain starts in state $s_i$, and let t be the column vector whose $i^{th}$ entry is $t_i$. The column vector of  $t$ can be written as follows:

\begin{equation}\label{eqn13}
t=\left[\begin{matrix}t({NA}_1)&t({NA}_2)&\begin{matrix}\ldots&t({NA}_n)\\\end{matrix}\\\end{matrix}\right]^T=NC
\end{equation}

where C is a column vector whose elements are one. Having calculated $t$, each of its elements represents the mean time to failure (MTTF) of the corresponding state.

In the area of reliability evaluation of UAV using Markov processes, there are a few existing research works. For example, Guo J. et al. \cite{guo2019reliability} focused on the balancing issue of the propulsion system in multi-rotor UAVs and proposed a solution based on Markov chains. Aslansefat, K. et al. \cite{aslansefat2019markov} proposed a set of Markov models for different possible configurations in the propulsion system of multi-rotor UAVs, taking into account not just the balancing issue but also the controllability aspects which make the model more dynamic. 

In this paper, the Markov models proposed by \cite{aslansefat2019markov} are used for the propulsion system and as a CBE (discussed in Section \ref{methodology}). As one of the contributions in this paper, the Markov models are also extended to use 'Motor status' as a symptom and be able to re-calculate the reliability and MTTF during the mission. For instance, if the real-time monitoring and diagnosis unit in the robot detects a motor failure in motor status based on the available sensors that the robot has, the proposed approach will select the equivalent state in the SMP that represents the motor failure and re-calculate the probability of failure and MTTF from that state to the failure state.

\subsection{Reliability Modeling using Arrhenius Equation}
\label{sec_ArrEq}
\pdfbookmark[subsection]{Reliability Modeling using Arrhenius Equation}{sec_ArrEq}
The lifetime reliability of a processing unit has a strong correlation with its temperature \cite{ottavi2014dependable}. Moreover, a processor's temperature depends on the UAV's performance and utilization. To capture this interaction between reliability and temperature, the Arrhenius equation has been used. The Arrhenius equation is used to compute the MTTF acceleration factor $(AF)$ depending on the processor's actual and reference temperatures.  
\begin{equation}
AF=e^{{\frac{Ea}{k}}(\frac{1}{T_{r}}-\frac{1}{T_{a}})}    
\end{equation}

where $Ea$ is the activation energy in electron-volts, $k$ is Boltzmann's constant (8.617E-05), $T_{r}$ is the reference temperature and $T_{a}$ is the actual temperature. 

The acceleration factor $(AF)$ is then used by the MTTF model to evaluate the effects of temperature on the MTTF. The final MTTF of the processor is calculated using the following equation:
\begin{equation}
MTTF=\frac{MTTF_{ref}}{AF}  
\end{equation}

where $MTTF_{ref}$ is the reference MTTF, estimated at the reference temperature. $MTTF_{ref}$ is usually given by the system's designers.

\subsection{The Executable Digital Dependable Identity (EDDI)} 
\label{sec_eddi}
\pdfbookmark[subsection]{The Executable Digital Dependable Identity (EDDI)}{sec_eddi}
Despite existing standards and guidelines, there is a great deal of variation in how assurance of dependability attributes is realized and claimed for concrete systems. This makes it difficult for third parties like certification authorities to analyze and evaluate the assurance approach in general, and especially when the systems are to be open, adaptive, or autonomous, like platooning cars \cite{kabir2019runtime}. 

To overcome this issue, Digital Dependability Identities (DDI) were created \cite{schneider2015wap,armengaud2021ddi}. 
A DDI is a structured, modular, and hierarchical model of a system's dependability properties. An assurance case is at the heart of the DDI, arguing for the assurance of the appropriate dependability attributes and connecting all models and artefacts (e.g. requirements, assumptions, architectural models, dependability analyses, evidences) essential for the argumentation. A DDI is created and updated throughout the design process, issued when the component or system is launched, and then maintained during the component or system's lifespan. DDIs are utilized for the hierarchical integration of systems to "systems of systems" in the field, as well as the integration of components to systems during development.

An Executable Digital Dependability Identity (EDDI) is an extension of the DDI concept that is intended to be executable at runtime. It leverages the design-time dependability models stored in the DDI and augments them with event monitoring and diagnostic capabilities to provide real-time feedback on reliability, security, and safety issues, thereby supporting safe operation and dynamic dependability management. Importantly, EDDIs are intended to act cooperatively when applied within a distributed multi-robot or multi-agent system, enabling on-the-fly reconfiguration, communication, and adaptation. The idea is to support dynamic adaptive system assurance and dependability management through event monitoring, run-time diagnostics, risk prediction, and recovery planning.

Like DDIs, EDDIs are based on the Open Dependability Exchange metamodel \cite{ODE_github}. An EDDI generally consists of some higher-level ODE-based system models for diagnostics, capability (e.g. success trees) and risk prediction (e.g. fault trees, Bayesian networks) and lower-level models for event monitoring and reliability estimation (e.g. Markov models, Bayesian networks). Once connected to sensor data and other pertinent system information, the EDDI can use these models to perform calculations to provide feedback and recommendations to the host system. 


\section{Methodology}
\label{methodology}
\begin{figure}
    \centering
    \includegraphics[scale=0.335]{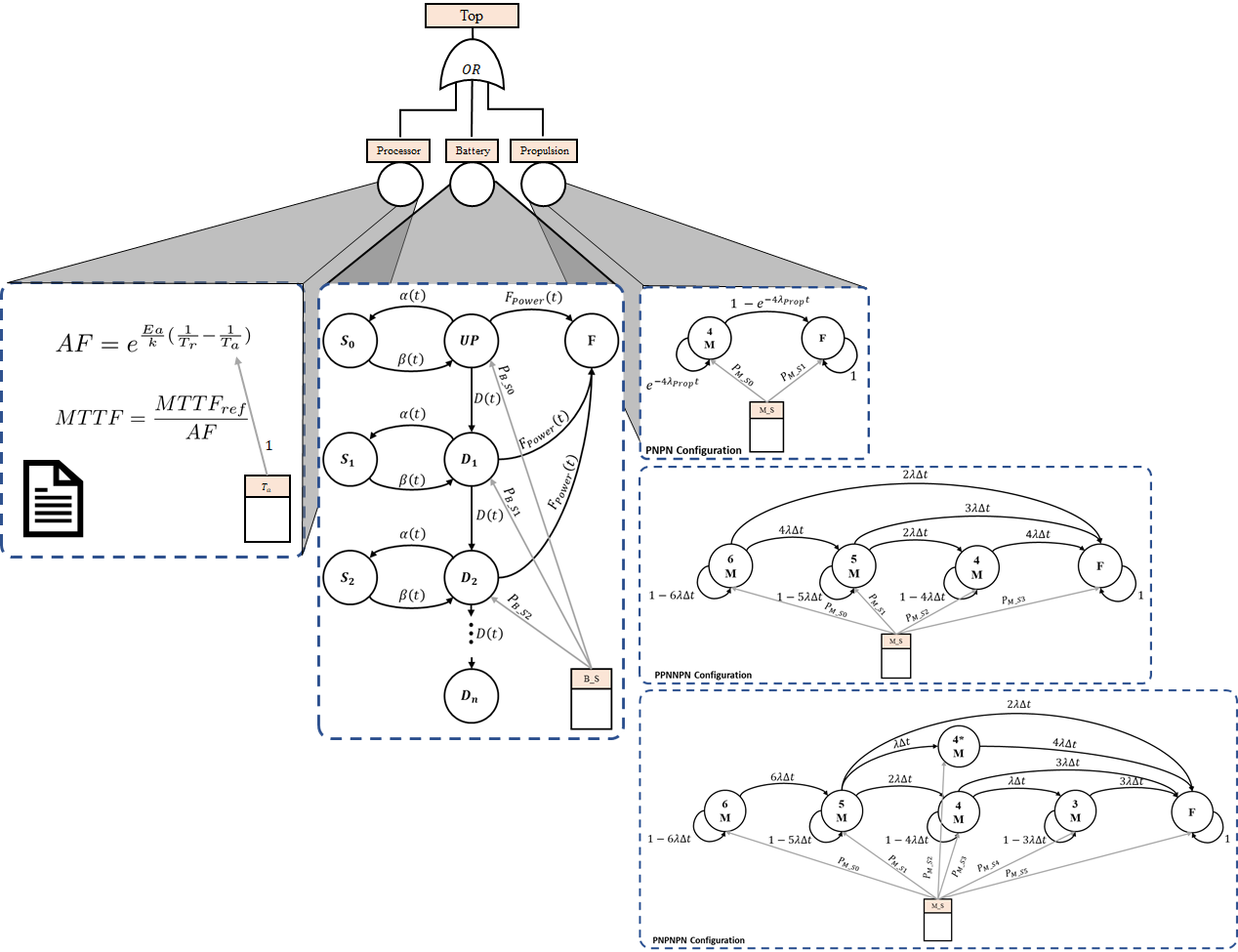}
    \caption{Small FTA of a UAV considering complex basic events with failure symptoms and three different types of propulsion system reconfiguration}
    \label{fig:FTA_CBE}
\end{figure}

\pdfbookmark[section]{Methodology}{methodology}
SafeDrones is an approach for real-time reliability and risk evaluation of multi-robot (multi-UAV) systems. The main goal of this work is to develop an early prototype instantiation of the EDDI concept for runtime reliability estimation for UAVs. It makes use of fault trees as the overall model with CBEs to support dynamic evaluation. A fault tree consisting of 9 main failure categories and 28 basic events is proposed for a generic  UAV in the appendix. However, to simplify the explanation of the methodology, a smaller fault tree of the UAV is provided in Figure \ref{fig:FTA_CBE}. 

The contribution and capabilities of the SafeDrones approach are as follows: 1) SafeDrones expands the idea of FTA with CBEs to not only consider SMPs but also other evaluation functions like the Arrhenius Equation; 2) it proposes the idea of having symptom events for each CBE; 3) it is also able to handle reliability evaluation of reconfigurable systems by using pre-defined models in one CBE (e.g. consider a hexacopter capable of reconfiguring its propulsion system on-the-fly from PNPNPN configuration to PPNNPN configuration), and 4) finally, SafeDrones provides Python functions which can be executed on each UAV and provide real-time reliability and MTTF evaluation. This paper primarily explores the first and fourth capabilities.

The tree provided in Figure \ref{fig:FTA_CBE} has three CBEs for battery failure, propulsion system failure and processor failure.
The processor failure has a symptom of actual temperature ($T_{a}$ is the symptom) and based on the Arrhenius Equation (see Section \ref{sec_ArrEq}), the reliability and the MTTF values of this basic event can be updated during the mission. The idea can be implemented for any component in the robot where its reliability can change based on temperature variation. The middle CBE is for the battery failure. This model is provided by \cite{Kim2010} and considers battery degradation as well as failure. In this paper, we have used the battery model with four degradation levels and the battery level status $B\_S$ is included as a symptom. So, based on the battery level status, the initial probability vector in the SMP will be updated and then the probability of failure (unreliability of the battery) will be updated accordingly. 

The third CBE is a propulsion system failure. The CBE is chosen to show the capabilities of  SafeDrones for handling system reconfiguration. The first configuration is for a quad-copter that has two propellers rotating clockwise (P) and two propellers rotating anti-clockwise (N) forming PNPN configuration. The second and the third configurations considered for hexa-copters with two different PNPNPN and PPNNPN configurations (P stands for clock wise rotation and N stands for anti-clockwise rotation). The detailed construction and simplification of these models has been discussed in our previous research ~\cite{aslansefat2019markov}.

\begin{figure}
    \centering
    \includegraphics[scale=0.38]{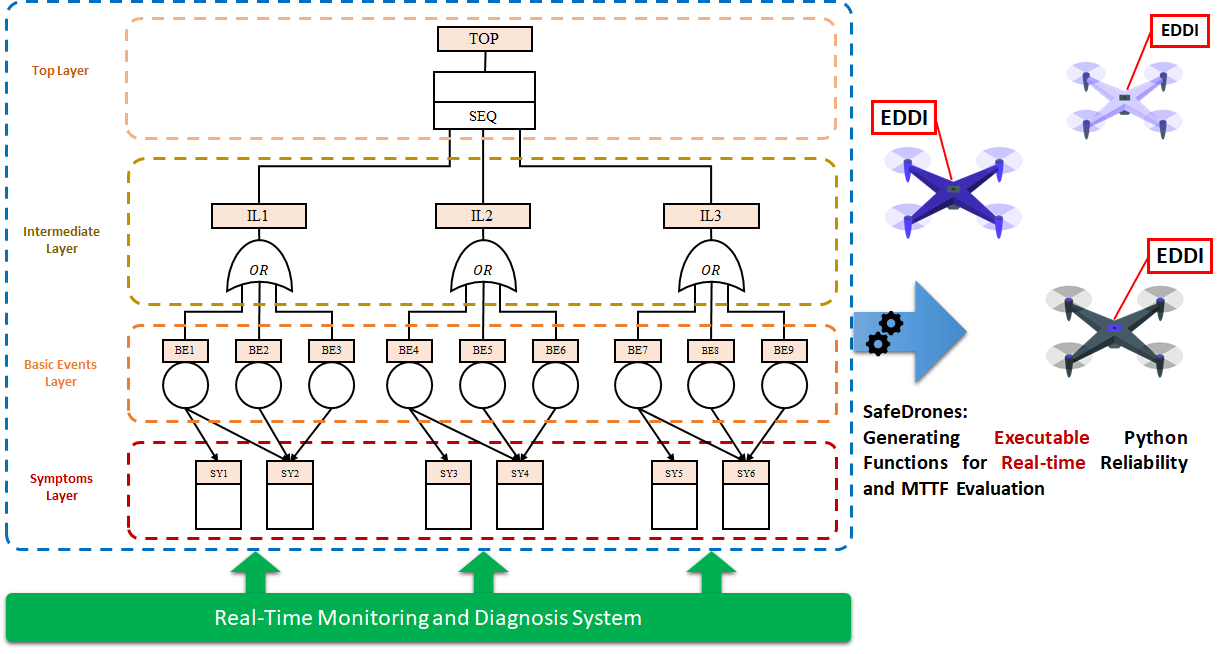}
    \caption{Overall view on merging real-time monitoring and diagnosis system with Fault Tree Analysis}
    \label{fig:FTA_2}
\end{figure}

Figure \ref{fig:FTA_2} further illustrates the idea of merging real-time monitoring and diagnosis with FTA. 
In a traditional FTA, the tree consists of a top layer, a number of intermediate layers, and a basic events layer. However, in our proposed approach there is a new layer called the symptoms layer. In the symptoms layer, the safety expert(s) should identify the potentially observable events in the system and define the relation between symptoms and basic events. For instance, in Figure \ref{fig:FTA_CBE}, the symptoms are temperature, battery status, and motor status along with motor configuration. In Figure \ref{fig:FTA_CBE}, it is assumed the temperature symptom only affects the processor and has no effect on the others. In this proposed reliability modeling approach, it is recommended to use CBEs to link with the symptoms. A CBE can take many forms, e.g. a multi-state Markov chain where the symptom affects its current state, a Bayesian Network where a symptom can form a belief, or some other reliability function where a symptom can be a parameter on it, etc. The link between symptoms and basic events can be both deterministic and probabilities values.

As discussed in Section \ref{sec_eddi}, the EDDI concept uses real-time evaluation of dependability attributes like reliability as function(s) to update the mission accordingly as part of a dependability-driven decision making system. This could lead to a variety of responses, such as reconfiguration during the mission (e.g. switching a hexacopter to quadcopter mode in the event of possible motor faults), changes to mission parameters (e.g. emergency landing or return to base sooner), or even requests for predictive maintenance of affected parts. In SafeDrones, all the calculations are implemented in Python (available in the GitHub repository) for runtime execution. The results could also be used by technologies like ConSerts \cite{schneider2010conditional} to generate conditional guarantee outcomes and provide the final decision accordingly. Moreover, based on the idea provided by \cite{Safety_AI}, it is possible to investigate the use of monitoring data to obtain safety model repair recommendations.

\section{Experimental Implementation}
\label{setup}
\begin{figure}[t]
    \centering
    \includegraphics[scale=0.4]{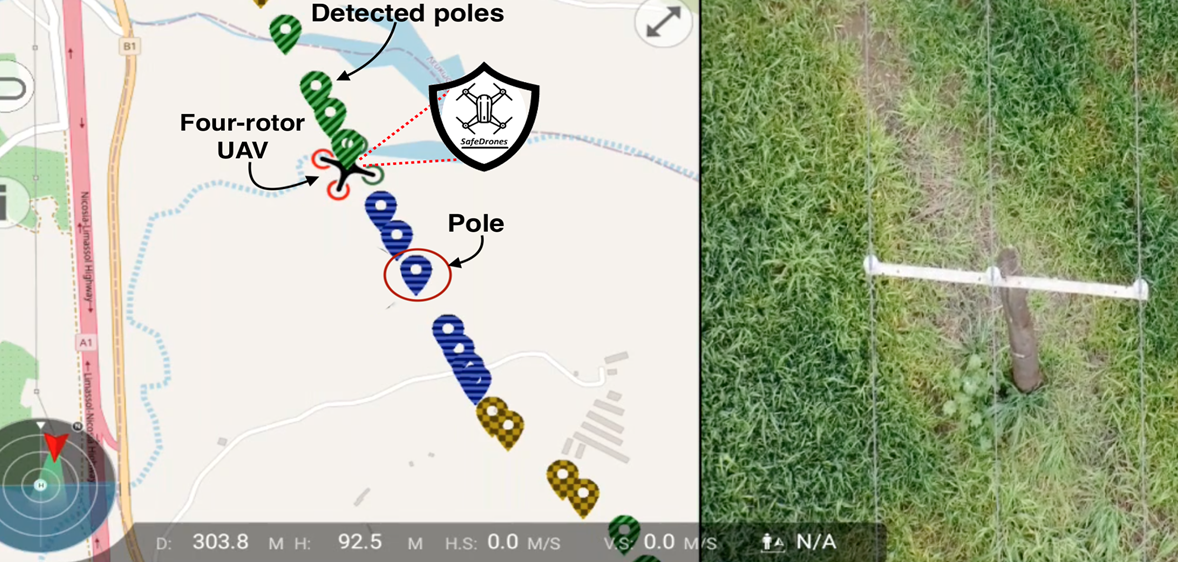}
    \caption{Inspection procedure using ICARUS toolkit \cite{savva2021icarus} for pole detection}
    \label{fig:poleDetection}
\end{figure}

\pdfbookmark[section]{Experimental Implementation}{setup}
To evaluate the reliability models presented in Section \ref{methodology} we use the ICARUS toolkit \cite{savva2021icarus}, which uses vision-based UAV monitoring platforms to automate the inspection of medium voltage power distribution networks. As Figure \ref{fig:poleDetection} shows, the UAV gathers data and provides a real-time data processing to identify poles and record their accurate positions. An off-the-shelf four-rotor UAV (DJI Matrice 300 RTK) equipped with different sensors, including temperature sensors, is used. On the top of the UAV, an NVIDIA Jetson Xavier NX embedded platform was mounted to run the deep learning and navigation algorithms, allowing the UAV to perform inspection procedures autonomously. Additionally, the UAV is equipped with the SafeDrones tool, which monitors parameters such as processor temperature, battery level and execution time to estimate UAV reliability. Furthermore, SafeDrones can recommend actions like mission abort and emergency landing if the estimated reliability falls below a predetermined threshold. For the pole detection task, the UAV flew at a constant height of approximately 50m above the ground with the camera turned downwards. To identify poles (top-view) in videos under different background and lighting conditions, the tiny-You-Only-Look-Once (tiny-YOLO) v4 was employed. 

For our analysis we monitor processor temperature and battery level every 1 second to estimate the probability of failure for the UAV using the models described earlier. All the other input parameters are shown in Table\ref{tab:inputparameters}. When the estimated probability of failure exceeds a specific threshold (we use 0.9 as a threshold for this analysis), an emergency action is taken. In this case, the action is to perform a safe emergency landing and continue the mission with another UAV. Note that the threshold can vary, depending on the mission and the time needed to safely land the UAV. The total execution time for the fault-free inspection mission to detect all the poles is 750 seconds.
\begin{table} [t]
\centering
\caption{Input values for the parameters used in the models, as  described in Sections \ref{Background} and \ref{methodology}}

\begin{tabular}{|p{0.15\linewidth} | p{0.3\linewidth}|p{0.5\linewidth}|} 
 \hline
 \centering
 Parameters & Description & Values \\ 
 \hline\hline
 \multicolumn{3}{||c||}{\textbf{Motor parameters}} \\
 \hline
 $MC$ & Motor Configuration & PNPN (P: positive clockwise direction, N: negative anti-clockwise direction) \\ 
 \hline
 Motor $\lambda$ & Motor failure rate & 0.001 \\
 \hline
 \multicolumn{3}{||c||}{\textbf{Battery parameters}} \\
  \hline
 Battery $\lambda$ & Battery failure rate & 0.0001 \\
 \hline
 $D$ & Battery degradation rate & 0.0064\\
\hline
 $\alpha$ & Battery usage rate & 0.008\\ 
 \hline
 $\beta$ & Battery inactivity rate & 0.007\\ 
 \hline
 \multicolumn{3}{||c||}{\textbf{Processor parameters}} \\
 \hline
 u & Utilization & 1\\
  \hline
 $MTTF_{ref}$ & Reference MTTF & 1000 hours\\
 \hline
 $E_{a}$ & Boltzmann constant & 8.617E-05\\
 \hline
 k & Activation energy & 0.3 electron-volts\\
 \hline
 $T_{r}$ & Reference temperature & 29$^{\circ}$ C \\
 \hline
\end{tabular}
 \label{tab:inputparameters}
\end{table}

To demonstrate the proposed concept, we use two scenarios:
\begin{enumerate}
\item	Fault-free scenario. In this scenario, all the components work properly without experiencing any faulty conditions.
\item	Faulty scenario. In this scenario, the battery stops working properly at a specific time $X$ causing a sharp drop in the battery level and at time $Y$, where $Y>X$, the processor starts overheating due to unexplained circumstances.
For this analysis, X equals to 250 seconds and Y equals to 400 seconds.

\end{enumerate}

\section{Experimental Results}
\label{results}
\pdfbookmark[section]{Experimental Results}{results}
This section reports the reliability analysis results for the two scenarios described in Section \ref{setup}.  

\subsection{Reliability Analysis of the Fault-Free Scenario}
\label{sec_raffs}
\pdfbookmark[subsection]{Reliability Analysis of the Fault-Free Scenario}{sec_raffs}

We first evaluate the probability of failure of the different components (battery and processor) and the total UAV for the fault-free scenario. It is assumed that the mission is about 800h. Figures \ref{fig:results} (a) and (b) show the battery level and processor temperature respectively (collected from UAV's telemetry logs), while Figures \ref{fig:results} (c) and (d) show the failure probability and MTTF for each component as well as the overall UAV. As Figure \ref{fig:results} (c) shows, the lower the battery level, the higher probability of failure. The sharp increase here when the battery level goes below 75\% is because our model discretizes the battery level into four states (25\% each), resulting in a jump when each discrete state is reached.  
Additionally, Figure \ref{fig:results} (c) shows that the processor's probability of failure is also related to the UAV's cumulative processing time. 
The exact correlation between reliability and processor temperature is shown in Figure \ref{fig:results} (i), which illustrates how the processor's MTTF changes according to the current temperature. As is is clearly shown, when the processor's temperature increases, the MTTF also decreases.  
Finally, as can be observed in Figure \ref{fig:results} (c), the overall UAV failure probability does not exceed the 0.90 threshold for emergency action, indicating that the inspection mission was completed successfully. Note that the threshold value should be determined by a team of safety experts.


\subsection{Reliability Analysis of the Faulty Scenario}
\label{sec_rafs}
\begin{figure}
    \centering
    \includegraphics[scale=0.46]{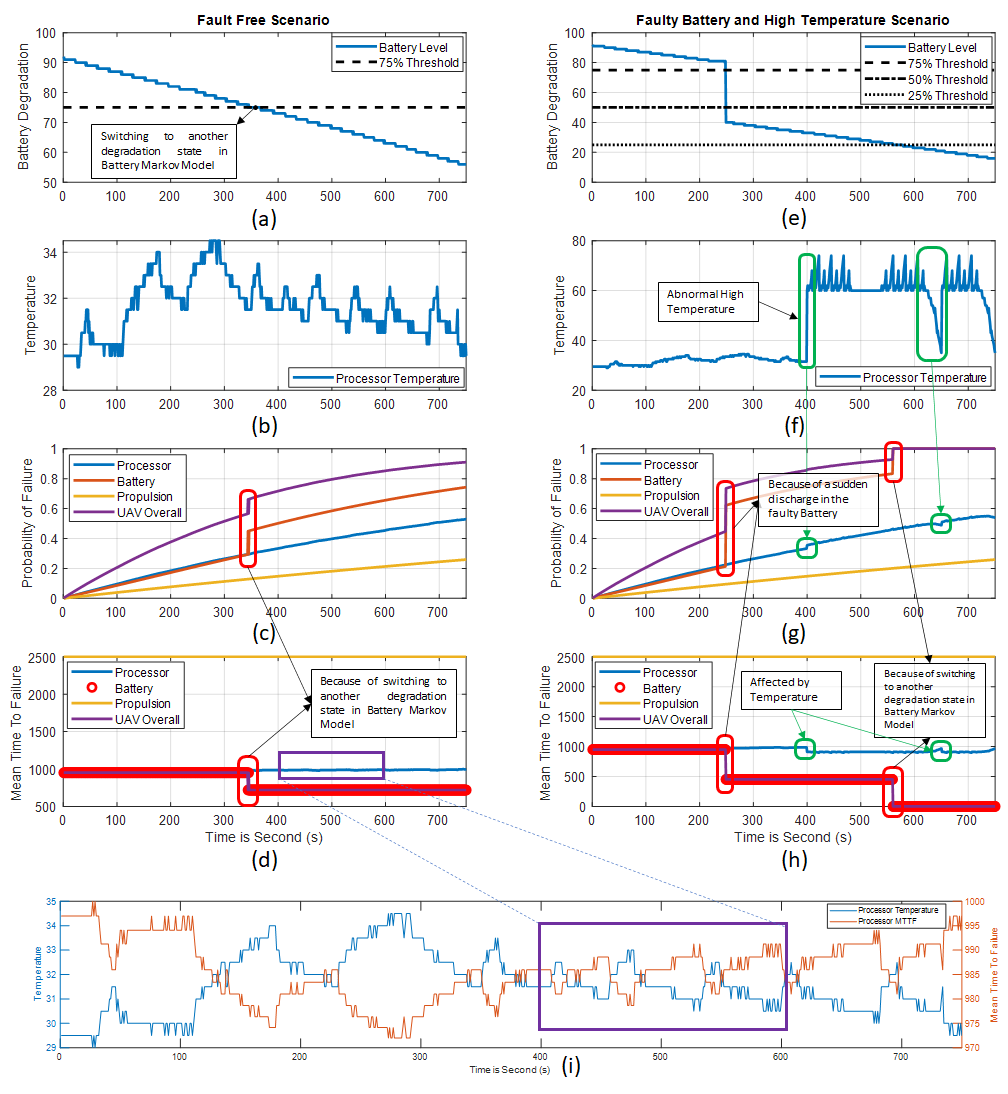}
    \caption{Fault-Free Scenario: (a) Battery degradation (battery level in percentage), (b) Processor Temperature (c) Probability of failure (d) Mean Time to failure -- Faulty Battery Scenario: (e) Battery degradation (battery level in percentage), (f) Processor Temperature (g) Probability of failure (h) Mean Time to failure -- (i) Processor’s MTTF and temperature for the Fault Free Scenario.}
    \label{fig:results}
\end{figure}

\pdfbookmark[subsection]{Reliability Analysis of the Faulty Scenario}{sec_rafs}
In the first scenario, the overall probability of failure was satisfactory throughout and the UAV managed to complete the mission safely. However, it is also possible for faults to develop in any component, and so in the second scenario we investigate how the probability of failure can be changed by simulating a faulty battery and the processor overheating. Here the battery stops working properly at the 250th second. At this point the battery level drops sharply from 80\% to 40\% as Figure \ref{fig:results} (e) shows. The processor's temperature also suddenly increases at the 400th second.
Figures \ref{fig:results} (g) and (h) depict the impact of these simulated faults on the probability of failure and MTTF respectively. As Figure \ref{fig:results} (g) shows the failure probability threshold of the UAV is exceeded at the 500th second. This leads to an emergency landing of the UAV even if the mission was not completed. In a multi-UAV scenario, another UAV can be dynamically tasked to continue and complete the mission in this case.

The results highlight the benefits of both the proposed SafeDrones approach and the overall EDDI concept in helping to avoid dangerous accidents caused by failures. By combining safety analysis models and reliability functions executable at runtime, we can obtain a more comprehensive overview of UAV dependability during real-time operation, one that takes into account multiple subsystems and sensors as well as predefined thresholds and corresponding mitigating actions. Such an approach is particularly valuable for autonomous platforms where there is no human operator to monitor safety directly.

\section{Conclusion and Future Work}
\label{sec_conclusion}
\pdfbookmark[section]{Conclusion and Future Work}{sec_conclusion}
To help address the problems of UAV reliability and risk assessment, particularly at runtime where operational and environmental factors are hard to predict, the SafeDrones reliability modeling approach has been proposed. It employs a combination of FTA with CBEs to support real-time reliability evaluation as a prototype of the EDDI concept. As part of this, it introduces a novel symptoms layer to integrate with runtime monitoring data. To illustrate SafeDrones, we applied it to a power network inspection use case to show how real-time reliability evaluation can be used to anticipate imminent failures and prevent accidents by recommending appropriate responses.






%
In this paper, we have focused on a single UAV. However, many UAV applications involve multiple UAVs. As part of our future work, we plan to extend SafeDrones to multiple UAVs to further explore the full EDDI concept and assess and preserve overall mission dependability in real-time. This demands modeling and dynamic evaluation of mission risk variability \cite{SINADRA.2020} and distributed dependability concept variability \cite{schneider2010conditional}, which can be realized e.g. by allowing dynamic task redistribution to the remaining UAVs if one or more UAVs have increased probability of failure. 

Furthermore, we plan to investigate other aspects of dependability by evaluating the reliability of machine learning components many UAVs have, e.g. for object detection. For this, we intend to make use of SafeML\cite{aslansefat2020safeml}\cite{aslansefat2021toward}. By using SafeML, we can obtain a more complete picture of real-time UAV dependability which can then be used to update the mission accordingly and improve mission completion. In the paper, it was assumed that the monitoring system is perfect, however, it would be possible to incorporate false positive, false negative and uncertainties as future work \cite{aslansefat2020performance}.

Finally, it is important to investigate how security issues will affect UAV operation \cite{valianti2021multi}. For this, we plan to consider a jamming attack scenario that will affect the communication with the GPS.


\section*{Code Availability}
To improve the research reproducibility, code, functions, demo notebooks, and other materials supporting this paper are published online at GitHub: 

\href{https://github.com/koo-ec/SafeDrones}{https://github.com/koo-ec/SafeDrones}.

\section*{Acknowledgement}
This work was supported by the Secure and Safe Multi-Robot Systems (SESAME) H2020 Project under Grant Agreement 101017258.

%
%
%
\bibliographystyle{splncs}

\begin{thebibliography}{10}
\providecommand{\url}[1]{\texttt{#1}}
\providecommand{\urlprefix}{URL }
\providecommand{\doi}[1]{https://doi.org/#1}

\bibitem{adler2007}
Adler, R., Forster, M., Trapp, M.: Determining configuration probabilities of
  safety-critical adaptive systems. In: 21st International Conference on
  Advanced Information Networking and Applications Workshops (AINAW'07).
  vol.~2, pp. 548--555. IEEE (2007)

\bibitem{armengaud2021ddi}
Armengaud, E., Schneider, D., Reich, J., Sorokos, I., Papadopoulos, Y., Zeller,
  M., Regan, G., Macher, G., Veledar, O., Thalmann, S., et~al.: Ddi: A novel
  technology and innovation model for dependable, collaborative and autonomous
  systems. In: 2021 Design, Automation \& Test in Europe Conference \&
  Exhibition (DATE). pp. 1626--1631. IEEE (2021)

\bibitem{aslansefat2020performance}
Aslansefat, K., Gogani, M.B., Kabir, S., Shoorehdeli, M.A., Yari, M.:
  Performance evaluation and design for variable threshold alarm systems
  through semi-markov process. ISA transactions  \textbf{97},  282--295 (2020)

\bibitem{aslansefat2021toward}
Aslansefat, K., Kabir, S., Abdullatif, A., Vasudevan, V., Papadopoulos, Y.:
  Toward improving confidence in autonomous vehicle software: A study on
  traffic sign recognition systems. Computer  \textbf{54}(8),  66--76 (2021)

\bibitem{aslansefat2020}
Aslansefat, K., Kabir, S., Gheraibia, Y., Papadopoulos, Y.: {Dynamic Fault Tree
  Analysis: State-of-the-Art in Modeling, Analysis, and Tools}. In: Reliability
  Management and Engineering: Challenges and Future Trends, chap.~4, pp.
  73--111. CRC Press- Taylor \& Francis (2020)

\bibitem{aslansefat2019hierarchical}
Aslansefat, K., Latif-Shabgahi, G.R.: A hierarchical approach for dynamic fault
  trees solution through semi-markov process. IEEE Transactions on Reliability
  \textbf{69}(3),  986--1003 (2019)

\bibitem{aslansefat2019markov}
Aslansefat, K., Marques, F., Mendon{\c{c}}a, R., Barata, J.: A markov
  process-based approach for reliability evaluation of the propulsion system in
  multi-rotor drones. In: Doctoral Conference on Computing, Electrical and
  Industrial Systems. pp. 91--98. Springer (2019)

\bibitem{aslansefat2020safeml}
Aslansefat, K., Sorokos, I., Whiting, D., Tavakoli~Kolagari, R., Papadopoulos,
  Y.: Safeml: Safety monitoring of machine learning classifiers through
  statistical difference measures. In: International Symposium on Model-Based
  Safety and Assessment. pp. 197--211. Springer (2020)

\bibitem{belcastro2017hazards}
Belcastro, C.M., Newman, R.L., Evans, J., Klyde, D.H., Barr, L.C., Ancel, E.:
  Hazards identification and analysis for unmanned aircraft system operations.
  In: 17th AIAA Aviation Technology, Integration, and Operations Conference.
  p.~3269 (2017)

\bibitem{bouissou2003}
Bouissou, M., Bon, J.L.: A new formalism that combines advantages of
  fault-trees and markov models: Boolean logic driven markov processes.
  Reliability Engineering \& System Safety  \textbf{82}(2),  149--163 (2003)

\bibitem{cochran2010wiley}
Cochran, J.: Wiley Encyclopedia of Operations Research and Management Science.
  John Wiley and Sons Ltd (2010)

\bibitem{ODE_github}
Consortium, D.: Open dependability exchange metamodel. {
  https://github.com/Digital-Dependability-Identities/ODE }, accessed:
  2022-04-28

\bibitem{franco2007failure}
Franco, B.J.d.O.M., G{\'o}es, L.C.S.: Failure analysis methods in unmanned
  aerial vehicle (uav) applications. In: Proceedings of COBEM 2007 19th
  International Congress of Mechanical Engineering (2007)

\bibitem{Safety_AI}
Gheraibia, Y., Kabir, S., Aslansefat, K., Sorokos, I., Papadopoulos, Y.:
  Safety+ ai: a novel approach to update safety models using artificial
  intelligence. IEEE Access  \textbf{7},  135855--135869 (2019)

\bibitem{guo2019reliability}
Guo, J., Elsayed, E.A.: Reliability of balanced multi-level unmanned aerial
  vehicles. Computers \& Operations Research  \textbf{106},  1--13 (2019)

\bibitem{Complex_BE}
Kabir, S., Aslansefat, K., Sorokos, I., Papadopoulos, Y., Gheraibia, Y.: A
  conceptual framework to incorporate complex basic events in hip-hops. In:
  International Symposium on Model-Based Safety and Assessment. pp. 109--124.
  Springer (2019)

\bibitem{kabir2019runtime}
Kabir, S., Sorokos, I., Aslansefat, K., Papadopoulos, Y., Gheraibia, Y., Reich,
  J., Saimler, M., Wei, R.: A runtime safety analysis concept for open adaptive
  systems. In: International Symposium on Model-Based Safety and Assessment.
  pp. 332--346. Springer (2019)

\bibitem{Kim2010}
{Kim}, D.S., {Ghosh}, R., {Trivedi}, K.S.: {A Hierarchical Model for
  Reliability Analysis of Sensor Networks}. In: 2010 IEEE 16th Pacific Rim
  International Symposium on Dependable Computing. pp. 247--248 (Dec 2010)

\bibitem{murtha2009evidence}
Murtha, J.F.: Evidence theory and fault tree analysis to cost-effectively
  improve reliability in small uav design. Virginia Polytechnic Inst. and State
  University  (2009)

\bibitem{olson2013qualitative}
Olson, I., Atkins, E.M.: Qualitative failure analysis for a small quadrotor
  unmanned aircraft system. In: AIAA Guidance, Navigation, and Control (GNC)
  Conference. p.~4761 (2013)

\bibitem{ottavi2014dependable}
Ottavi, M., Pontarelli, S., Gizopoulos, D., Bolchini, C., Michael, M.K.,
  Anghel, L., Tahoori, M., Paschalis, A., Reviriego, P., Bringmann, O., et~al.:
  Dependable multicore architectures at nanoscale: The view from europe. IEEE
  Design \& Test  \textbf{32}(2),  17--28 (2014)

\bibitem{SINADRA.2020}
Reich, J., Trapp, M.: {SINADRA: Towards a Framework for Assurable
  Situation-Aware Dynamic Risk Assessment of Autonomous Vehicles}. In: 16th
  European Dependable Computing Conference, {EDCC} 2020, Munich, Germany,
  September 7-10, 2020. pp. 47--50. {IEEE} (2020).
  \doi{10.1109/EDCC51268.2020.00017}

\bibitem{sadeghzadeh2011fault}
Sadeghzadeh, I., Mehta, A., Zhang, Y.: Fault/damage tolerant control of a
  quadrotor helicopter uav using model reference adaptive control and
  gain-scheduled pid. In: AIAA Guidance, Navigation, and Control Conference.
  p.~6716 (2011)

\bibitem{savva2021icarus}
Savva, A., Zacharia, A., Makrigiorgis, R., Anastasiou, A., Kyrkou, C., Kolios,
  P., Panayiotou, C., Theocharides, T.: Icarus: Automatic autonomous power
  infrastructure inspection with uavs. In: 2021 International Conference on
  Unmanned Aircraft Systems (ICUAS). pp. 918--926. IEEE (2021)

\bibitem{schneider2010conditional}
Schneider, D., Trapp, M.: Conditional safety certification of open adaptive
  systems. ACM Trans. Auton. Adapt. Syst.  \textbf{8}(2) (jul 2013).
  \doi{10.1145/2491465.2491467}

\bibitem{schneider2015wap}
Schneider, D., Trapp, M., Papadopoulos, Y., Armengaud, E., Zeller, M.,
  H{\"o}fig, K.: Wap: digital dependability identities. In: 2015 IEEE 26th
  International Symposium on Software Reliability Engineering (ISSRE). pp.
  324--329. IEEE (2015)

\bibitem{sharvia2016}
Sharvia, S., Kabir, S., Walker, M., Papadopoulos, Y.: Model-based dependability
  analysis: state-of-the-art, challenges, and future outlook. In: Software
  Quality Assurance, pp. 251--278. Elsevier (2016)

\bibitem{Amazon_Air_Bloomberg}
Soper, S., Day, M.: Amazon drone crashes hit jeff bezos’ delivery dreams.
  {
  https://www.bloomberg.com/news/features/2022-04-10/amazon-drone-crashes-delays-put-bezos-s-delivery-dream-at-risk
  }, accessed: 2022-04-10

\bibitem{trivedi2017reliability}
Trivedi, K.S., Bobbio, A.: Reliability and availability engineering: modeling,
  analysis, and applications. Cambridge University Press (2017)

\bibitem{valianti2021multi}
Valianti, P., Papaioannou, S., Kolios, P., Ellinas, G.: Multi-agent coordinated
  close-in jamming for disabling a rogue drone. IEEE Transactions on Mobile
  Computing  (2021)

\bibitem{Vesely2002}
Vesely, W., Dugan, J., Fragola, J., Minarick, Railsback, J.: {Fault Tree
  Handbook with Aerospace Applications}. Tech. rep., N{ASA} office of safety
  and mission assurance, Washington, DC (2002)

\end{thebibliography}

%





\newpage
\section*{Appendix}


\subsection{Proposed Fault Tree of a generic UAV}
\begin{figure}
    \centering
    \includegraphics[scale=0.35]{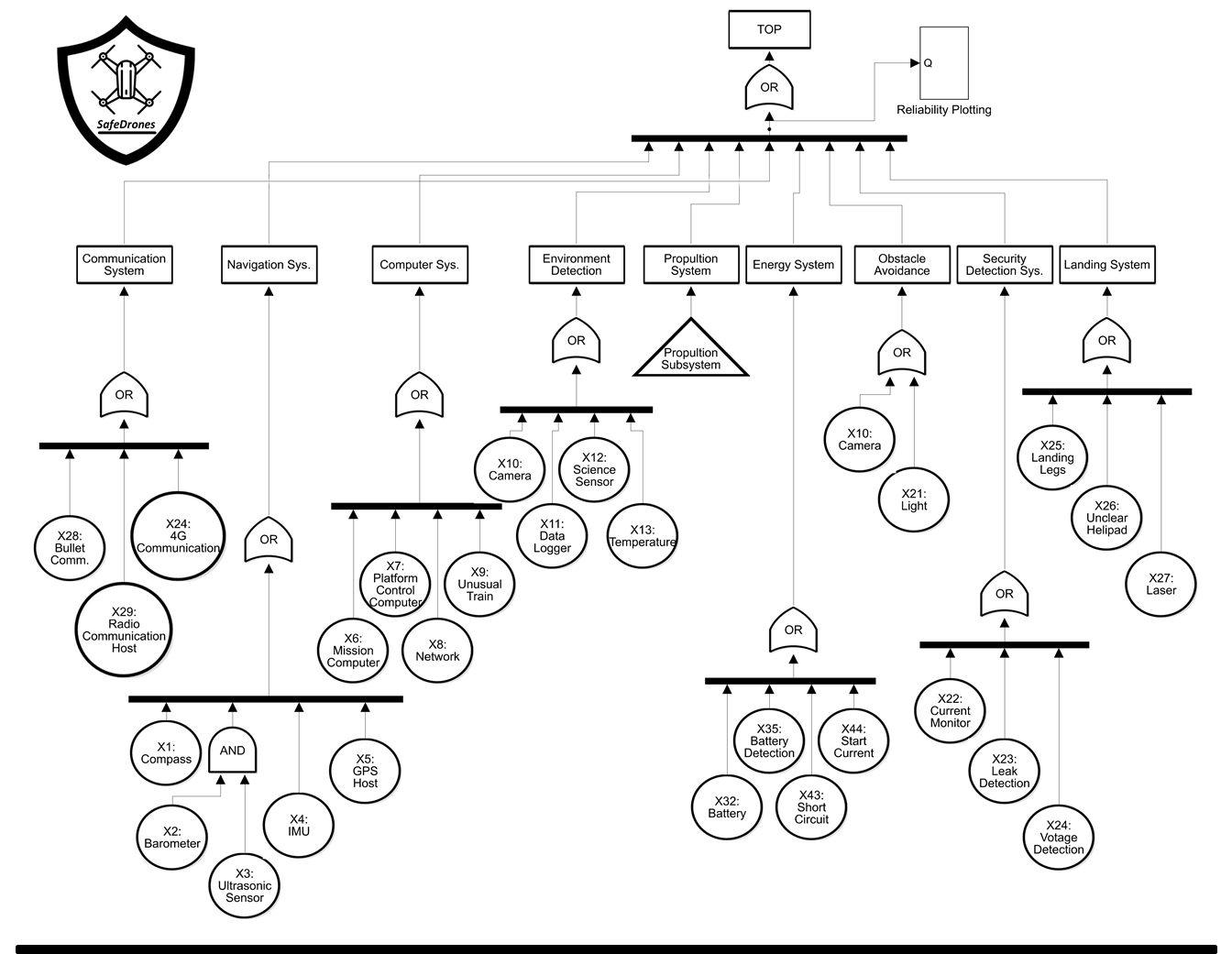}
    \caption{Proposed Fault Tree of a generic UAV}
    \label{fig:Full_FTA}
\end{figure}

Figure \ref{fig:Full_FTA} illustrates the proposed Fault Tree of a generic UAV consist of nine failure categories including: I) Communication system failure, II) navigation system failure, III) Computer system failure, IV) Environment detection systems, V) Propulsion system, VI) Energy system, VII) Obstacle avoidance system, VIII) Security system, and IX) Landing system.

\end{document}